\documentclass[12pt]{article}

\usepackage{scicite}
\usepackage{amsmath}
\usepackage{graphicx}
\usepackage{dcolumn}
\usepackage{bm}
\usepackage{times}
\DeclareMathOperator{\sign}{sign}

\newenvironment{sciabstract}{%
\begin{quote} \bf}
{\end{quote}}

\title{Multi-legged matter transport: a framework for locomotion on noisy landscapes}

\author
{Baxi Chong$^{1,2}$, Juntao He$^{3}$, Daniel Soto$^{3}$, Tianyu Wang$^{3}$, \\ Daniel Irvine$^{4}$, Grigoriy Blekherman$^{4}$, Daniel I. Goldman$^{1,2,3,\ast}$\\
\\
\normalsize{$^{1}$Interdisciplinary Graduate Program in Quantitative Biosciences, Georgia Institute of Technology,}\\
\normalsize{North Avenue, Atlanta, GA 30332, USA}\\
\normalsize{$^{2}$School of Physics, Georgia Institute of Technology,}\\
\normalsize{837 State St NW, Atlanta, GA 30332, USA}\\
\normalsize{$^{3}$Institute for Robotics and Intelligent Machines, Georgia Institute of Technology,}\\
\normalsize{801 Atlantic Dr NW, Atlanta, GA 30332, USA}\\
\normalsize{$^{4}$School of Mathematics, Georgia Institute of Technology,}\\
\normalsize{686 Cherry St NW, Atlanta, GA 30332, USA}\\
\\
\normalsize{$^\ast$To whom correspondence should be addressed; E-mail: daniel.goldman@physics.gatech.edu}
}

\date{}

\begin{document} 


\baselineskip24pt


\maketitle


\begin{sciabstract}
While the transport of matter by wheeled vehicles or legged robots can be guaranteed in engineered landscapes like roads or rails, locomotion prediction in complex environments like collapsed buildings or crop fields remains challenging. Inspired by principles of information transmission which allow signals to be reliably transmitted over noisy channels, we develop a ``matter transport" framework demonstrating that non-inertial locomotion can be provably generated over ``noisy" rugose landscapes (heterogeneities  on the scale of locomotor dimensions). Experiments confirm that sufficient spatial redundancy in the form of serially-connected legged robots leads to reliable transport on such terrain without requiring sensing and control. Further analogies from communication theory coupled to advances in gaits (coding) and sensor-based feedback control (error detection/correction) can lead to agile locomotion in complex terradynamic regimes.
\end{sciabstract}


\section*{Introduction}

The transport of matter across land is crucial to societies and groups~\cite{aguilar2018collective,mindell2011transportation,kreutzmann1991karakoram}, as well as individual locomotors~\cite{inman1966human}. Engineered self-propulsion as a means of terrestrial matter transport has been studied across scales from enormous multi-wheeled trains to small few-wheeled and legged robots~\cite{holmes2006dynamics,raibert2008bigdog,saranli2001rhex}. For large devices, low-dissipation, inertia-dominated locomotion is a commonly used matter transport scheme. Specifically, locomotion in wheeled systems on smooth surfaces like tracks and roads will persist over long distance unless acted on by dissipative internal or external forces (Fig.~1.A).

On natural terrain, dissipation due to external forces can occur via interactions with terrain heterogeneities such as obstacles, gaps, or inclines~\cite{bekker1960off} as well as via interactions with flowable materials~\cite{li2013terradynamics}. In dissipation-dominated applications, such as those encountered in robot movement in certain agricultural (e.g., crop fields) or confined and crowded search-and-rescue (e.g., collapsed buildings) scenarios, a system must continuously and actively generate forces and/or reduce dissipation. In part, because of our lack of understanding of the terradynamic~\cite{li2013terradynamics,li2015terradynamically} interactions with the environments listed above, principles by which locomotors can be designed and controlled to guarantee reliable/predictable matter transport are lacking.

One engineering solution~\cite{weber1836mechanik} to facilitate matter transport in complex terradynamic regimes is to use structures like limbs to periodically make and break contact with the environment~\cite{lee2020learning}. Such dynamics can potentially simplify the thrusting interactions into a collection of discrete units which minimize unexpected interference~\cite{holmes2006dynamics}, thus providing an alternative to wheeled carriers on ``noisy" landscapes. There have been two basic approaches of limb use in dissipation-dominated environments. The first relies heavily on sensors~\cite{borenstein1997mobile} to detect and respond to terrain heterogeneity in real-time~\cite{raibert2008bigdog,hutter2016anymal}. This approach is used for the increasingly agile locomotion in state-of-the-art legged robots (mostly bipedal or quadrupedal)~\cite{westervelt2018feedback,raibert2008bigdog,hutter2016anymal,lee2020learning}. However, the use of sensors and high bandwidth control can be expensive and restricted to specific applications. 

The second approach has been to instill legged robots with ``mechanical intelligence" such that locomotion can be performed with minimal environmental awareness. This has been most effective with devices with more than four legs, e.g., hexapods~\cite{saranli2001rhex} and myriapods~\cite{ozkan2021self,aoi2016advantage}. While more limbs help avoid catastrophic failures (e.g., loss of stability), terrain heterogeneity can still cause deficiencies in thrusting interactions which significantly reduce locomotor performance (Movie S1 and~\cite{moore2002reliable,ordonez2013terrain,wu2016integrated}). This raises the question as to how variable numbers of limbs and sensors should be arranged such that one can guarantee that a locomotor can go from A to B in a specified time across an arbitrarily complex landscape and further bound how much sensing/feedback/bandwidth/control is needed.

This question is analogous to that of information and signal transmission over noisy channels as first analyzed by Shannon nearly a century ago~\cite{shannon1948mathematical}. Over a noiseless channel, a continuous analog signal is, in principle, able to convey an infinite amount of information~\cite{feinstein1954new}. Despite its efficiency, the analog signal can be distorted by channel noise inherent in all communication modalities, a property similar to inertia-driven matter transport. To counter channel noise, Shannon~\cite{shannon1948mathematical} constructed a scheme, in which the central idea was to digitize (encode) information into binary bit sequences and ``buffer" (correct) the transmission error via redundancy (Fig.~1.C). 

With the analogy to information theory, it is reasonable, at least conceptually, to anticipate reliable matter transport over noisy landscapes given sufficient redundancy of terrestrial interaction. Here we illustrate that this anticipation is correct and leads to an open-loop framework for matter transport by which, for a complex terradynamic task, we can guarantee that sufficiently redundant multi-legged robots can reliably and predictably self-transport over some distance via ``buffer and tolerate" dynamics without need for sensing and feedback control or environmental awareness (Fig.~1.B, Movie S1 and~\cite{ozkan2021self}). 

\section*{Development of the matter transport framework}

Our framework proceeds as follows (Fig.~2): we define a transport task as a physical entity moving to a specific destination $D$ at a fixed time $T$. As shown in Fig.~2.B.I, this is analogous to the intended message being transmitted across a noisy channel in a given rate. Similar to the bit-based digital signal transmission, we focus on the dissipation-dominated system where locomotion is driven by thrusting interactions from basic active contacts (bacs, our analogy to bits), discrete units of active terradynamic interaction. Examples of bacs include limbs making contact with the environment~\cite{full1999templates} or vertical waves of contact in limbless robots~\cite{astley2015modulation}. We quantify the temporal and spatial distribution of bacs as a binary sequence $X^{m}$, where 1 denotes contact and 0 denotes non-contact (Fig.~2.B.II). As the locomotor implements the desired bac sequence over a noisy landscape (the analogue of a noisy channel, Fig.~2.B.III), the terrain uncertainty can introduce contact noise to the actual bac sequence, $Y^{m}$ (Fig.~2.B.IV). This leads to a discrepancy between the actual destination $\hat{D}$ (evaluated at the scheduled time $T$) and the desired destination, $D$ (Fig.~2.B.V).

We next discuss our characterization and quantification of noisy landscapes. A dissipation-dominated terrain can have different types of heterogeneities, each with different complex terradynamic effects on bacs~\cite{li2013terradynamics}. Consider a terrain characterized by a height map, $h(x,y)$. Depending on the scale of the gradient, $[\partial h/\partial x,\ \partial h/\partial y]$, the terrain heterogeneity can affect the locomotion in the form of slopes, walls, or obstacles (Fig~1.B), which directly impact the thrust-generation process in the plane parallel to the terrestrial surface (e.g., a stumble~\cite{forssberg1979stumbling}). Parallel thrust disturbances can be minimized by proper design of mechanical structures or passively compliant mechanisms (see SM Section 1.2 and~\cite{ozkan2021self,travers2018shape}). Here, we focus on a class of noisy landscapes (rugose terrains) where the height distribution, $h(x,y)$, can affect the supporting force distribution (e.g., missing steps) in the direction perpendicular to the terrestrial plane and therefore contaminate the intended bac sequence ($X^m\rightarrow Y^m$). 

With the notion of bacs and contact noise established, we can now model matter transport as a stochastic process. We first consider an abstract characterization of thrust generation given one pair of legs. We quantify the instantaneous thrust over a bac, $f(t)$, as the instantaneous external force required to keep the locomotor in place at time $t\in[0,\ \tau)$, where $\tau$ is the duration of the bac. An example of thrust function $f(t)$ is illustrated in Fig.~3.A. The nominal (undisturbed on flat terrain) average thrust is: $f_n = \frac{1}{\tau}\int_0^{\tau} f(t)dt$. 

Next we introduce a coefficient function which encapsulates the uncertainty in the bac, $c(t)$. The terrain-disturbed thrust can be formulated by $\hat{f} =\frac{1}{\tau} \int_0^{\tau} c(t)f(t)dt$. We assume $c(t)$ has the property $\frac{1}{\tau} \int_0^{\tau} c(t) = 1$, so that the supporting force balances gravity. Further, we assume that the initiation of a bac is delayed by some time $c_1$, and the duration of a bac is shortened to $\tau_u$: $\{c(t) = 0, t\not\in[c_1,\ c_1+\tau_u]\}$. Specifically, $c_1$ is assumed to be a random variable from a uniform distribution $c_1\sim U(0,\tau)$, and the duration of the bac, $\tau_u$, is assumed to be a random variable determined by the terrain rugosity. We sample $\tau_u$ from the cumulative distribution function given by $G(\tau_u)=(1-b)\tau_u/\tau+b$, $\tau_u\in [0,\ \tau]$ so that there is a finite probability of complete bac loss: $p(\tau_u = 0)=b$, and $b<1$ characterizes the contact noise level and offers an approximation to the rugosity of the terrain (Fig.~3.C). Where $c_1+\tau_u>\tau$, we extend the excessive contact duration ($c_1+\tau_u-\tau$) into the next bac (Fig.~2.B.III and SM Section 1.4). For simplicity, we assume that $c(t)$ is otherwise uniformly distributed during the bac: $\{c(t) = \tau_u^{-1}\tau, t\in[c_1,\ c_1+\tau_u]\}$. In this way, the terrain-disturbed average thrust reduces to: $\hat{f} = \sign(\tau_u)\tau_u^{-1} f_u$, where $f_u = \int_{c_1}^{c_1+\tau_u} f(t)dt$ is the thrust disturbance. The sign function $\sign(\tau_u)$ implies that no thrust will be generated ($\hat{f}=0$) with complete bac loss ($\tau_u = 0$).

As suggested in previous studies of dissipation-dominated multi-legged locomotion~\cite{chong2023self,zhao2022walking}, because of the periodic limb lifting and landing, an effective viscous (rate-dependent) cycle-averaged thrust-velocity (the average thrust and velocity over a period respectively) relationship emerges in frictional environments, despite such thrusts being instantaneously independent of velocity. Specifically, the relationship of cycle-averaged robot locomotion velocity is derived to be linearly correlated with the cycle-averaged thrust: $\hat{v}=\gamma^{-1}\hat{f}$, where $\gamma$ is the effective-viscous drag coefficient (Fig.~3.D). In this way, the terrain-disturbed velocity can be approximated by $\hat{v} = \gamma^{-1}\sign(\tau_u)\tau_u^{-1} f_u$.

Taking the analogy from information theory in which redundant bits can bound the uncertainty from channel noise, we hypothesize that locomotors with redundant bacs can offer robustness over terrain uncertainty. A straightforward scheme to include redundancy is to decrease the transport rate by allowing more transport time (temporal redundancy). Thus, we have:

\begin{equation}
    \hat{v}^{[1]}_T = \frac{1}{\gamma T}\sum_{i=1}^{T}  \sign(\tau_u^i)\frac{f_u^i}{\tau_u^i},
\end{equation}

\noindent where $\hat{v}^{[1]}_T$ is the average terrain-disturbed velocity over $T$ periods, $\tau_u^{i}$ and $f_u^{i}$ are the contact and thrust disturbance, respectively, over the $i$-th period. $T$ here represents the order of temporal redundancy. We expect the variance of the average terrain-disturbed velocity, $\sigma^2(\hat{v}^{[1]}_T)$, to decrease as $T$ increases. Further, $\hat{v}^{[1]}_T$ converges to a Dirac delta function as $T$ approaches infinity (proof given in the SM, Proposition 3). Moreover, the expected average terrain-disturbed velocity, $\langle\hat{v}^{[1]}_T\rangle$, remains constant (law of large numbers). 

We next evaluate the effectiveness of this temporal redundancy scheme. For simplicity, we assume $D$ is one-dimensional, $\hat{D}^{[1]}_T=T\hat{v}^{[1]}_T$, and we note that there is no reason to expect that $\hat{D}^{[1]}_T$ should converge as $T$ increases. 
Therefore, temporal redundancy can only guarantee the completion of a matter transport task, but the exact transport duration can be indefinite. 

Given the inefficiency of temporal redundancy, we develop a framework, in analogy to Shannon's encoding scheme, to remove inefficient redundancy and compensate it with ``redundancy of the right sort"~\cite{pierce2012introduction}[p.~164] for more effective locomotion. 
In particular, the appropriate redundancy facilitates the simultaneous ``communication" (e.g., re-distribution) of bacs in response to contact noise. 
To develop a specific scenario for legged systems, we consider redundancy in the form of repeating serially-connected modules, where a module is defined as a pair of legs. With proper coordination, the effect of contact noise will be shared by all bacs instead of acting on an individual bac. Because such redundancy is distributed in the spatial domain, we refer to it as spatial redundancy. Effectively, this spatial redundancy serves as a moving average filter over the contact noise profile. For simplicity, we consider a simple module coordination that the instantaneous thrust $f(t)$ on each module is identical and invariant to the number of modules. The average terrain-disturbed velocity for $N$ serially connected modules over $T$ periods is:

\begin{equation}
    \hat{v}^{[N]}_T = \frac{1}{\gamma T} \sum_{i = 1}^{T}\Bigg(\sign(\sum_{j=1}^{N}  \tau_u^{ij})\frac{\sum_{j=1}^{N} f_u^{ij}}{\sum_{j=1}^{N}  \tau_u^{ij}} \Bigg),
\end{equation}

\noindent where $\tau_u^{ij}$ and $f_u^{ij}$ are disturbances on the $j$-th module over the $i$-th temporal repetition. Intuitively, in the case where there are $M$ complete bac losses in the $i$-th temporal repetition, $M = |\{j, \tau_u^{ij}=0\}|$, the locomotor with $N$ modules will essentially reduce to the configuration with $N-M$ modules. In other words, locomotors with spatial redundancy $N$ can afford up to $N-1$ complete bac losses without significant thrust deficiency, indicating that spatial redundancy can also serve to bound the uncertainty in thrust generation. We show that for fixed $T\geq1$, $\hat{v}^{[N]}_T$ will also converge to a Dirac delta function as spatial redundancy $N$ approaches infinity (proof given in the SM, Proposition 4). The expected average terrain-disturbed velocity, $\langle\hat{v}^{[N]}_T\rangle$, can be approximated by $(1-b^N)C_{s}$ where $C_{s}$ is a constant determined by $f(t)$, $\gamma$, and $b$ (proof given in the SM, Proposition 4). Therefore, greater spatial redundancy not only reduces variance but also improves the expected average terrain-disturbed velocity, a feature otherwise not possible with only temporal redundancy. 

For any fixed $T$, the distribution of actual destinations $\hat{D}^{[N]}_T=T\hat{v}^{[N]}_T$ will also converge to a Dirac delta function as $N$ approaches infinity. Given a matter transport task over desired the distance $D$ at the scheduled time $T$, we consider it successful if 
$$\vert\vert (\hat{D},\hat{T}) - (D,T) \vert\vert= |\hat{D}^{[N]}_T-D|<\epsilon,$$ where $\epsilon$ is the tolerance. In this way, for arbitrary $\epsilon>0$ and $p_0<1$, there exists a finite $N$ such that the probability of successful matter transport (subject to the tolerance $\epsilon$) is greater than $p_0$ (proof given in the SM, Proposition 5). The ``minimal spatial redundancy" required to achieve the probability of $p_0$ for tolerance $\epsilon>0$ is bounded by: $N_{\epsilon,p_0} \leq \frac{\log(1-\sqrt[k D]{p_0})}{\log (b)}$, where $k $ is a constant describing the locomotor speed on flat terrain, and $D$ is the desired destination specified by distance (proof given in the SM, Proposition 5 and 6).

\section*{Numerical tests of the framework}

We first tested our theoretical predictions by numerical simulations. We chose a bac spatio-temporal distribution pattern based on limb stepping patterns of biological centipedes (and whose efficacy in generating locomotion in multi-legged robots was previously studied in~\cite{chong2022general, chong2023self}). Each bac generates an instantaneous thrust given by $f(t)$ which is independent of our choice of spatial redundancy (proof given in the SM, Proposition 1). We illustrate $f(t)$ in Fig.~3.B. Assuming $b=0.5$, we compare the distribution of normalized (by the nominal speed $v_{\text{open}}$) terrain-disturbed velocity, $v/v_{\text{open}}$, for different combinations of temporal and spatial redundancy in Fig.~3.E. In this example, $C_{s}=v_{\text{open}}$, indicates that the terrain-disturbed velocity converges to the nominal velocity (dashed black curve) given sufficient spatial redundancy. 

To illustrate that spatial redundancy allows the matter transport process (defined by the actual distance achieved, $\hat{D}$, and duration to achieve this displacement, $\hat{T}$) to converge to the desired transport distance and duration pair $(D,T)$, we calculate the 90\% CI of $\hat{D}$ as a function of $T$  for $N=8$ (red curves) and $N=1$ (blue curves). $T$ is plotted against $\hat{D}$ in Fig.~3.F. With greater spatial redundancy, the variation of $\hat{D}$ is significantly reduced, approaching the limit of nominal distance-duration relationship (dashed black curve). Further, we numerically calculate the probability of successful matter transport (evaluated at $T=1$) as a function of spatial redundancy $N$ subject to different choices of tolerance $\epsilon$ (Fig.~3.G). For all choices of $\epsilon$, the success probability converges to one as $N$ increases. Fig.~3.G also indicates that the marginal benefit of having more legs decreases as $N$ increases; the predicted limit is given by the dashed black curve. Finally, the locomotion performance can be affected by the terrain rugosity. We calculated $N_{\epsilon, p_0}$  as a function of $b$, the contact noise level and plot this in Fig.~3.H. We compare three combinations of $\epsilon$ (tolerance) and $p_0$ (success probability), and notice more rugose terrains require greater (yet finite) $N_{\epsilon, p_0}$; the theoretical bound is given by the dashed black curve when $p_0=0.9$.

\section*{Experimental tests of the framework}

Because the model assumes simplified environmental interactions, we next tested if our framework could predict locomotor performance in the physical world. We chose to work with a well-controlled laboratory multi-legged robophysical model (movie S1, design and control details are provided in the SM, Section 1.2 and 1.3), similar in design to that in~\cite{ozkan2020systematic,chong2022general,chong2023self}, measuring the average speed and its variation in such robots with different leg numbers and terrain complexity. Importantly, given that our framework indicates that matter transport can be achieved without need for environmental awareness, all robots were controlled such that they executed their pre-programmed stepping patterns (open-loop) and did not sense or respond to features of the environment. To facilitate comparison across different spatial redundancy $N$, our chosen bac sequence (the same as that in the numerical tests) has the property that all robots share the same thrust function $f(t)$, the same performance on flat terrain, and the same thrust-velocity relationship (proof given in the SM, Proposition 2). 

We constructed laboratory models of rugose terrains composed of ($10\times10$ cm$^2$) blocks with variation in heights (Fig.~4.A). The block heights, $h(x,y)$, are randomly distributed (SM, Section 1.1). Such rugose terrains ensure that limbs will experience thrust deficiency from stochastic contact~\cite{jacoff2008stepfield}. Note that the contact error can also arise from robot motor noise due to actuation delay, insufficient torque, or body compliance. We define the terrain rugosity, $R_g$, as the standard deviation of heights normalized by block side length. We tested the performance of 3-8 segmented (with 6-16 directionally-compliant limbs~\cite{ozkan2020systematic}) robophysical models on rugose terrains and recorded the bac duration ($\tau_u$) on each leg. The distributions of $\tau_u$ measured from 225 and 309 bacs for terrain with rugosity 0.17 and 0.32, respectively, are shown in Fig. S5. For $0<\tau_u<\tau$, the measured cumulative distribution function of $\tau_u$ can be approximated by a linear function, a feature in accord with (and therefore, justifying) our assumed bac duration distribution in Fig.~3.B. 

We also recorded the normalized inter-step average velocity (the average velocity over a step spanning the duration of a bac) of the 12-legged robot on rugose terrains during multiple steps ($v/v_{\text{open}}$) and compared this to the corresponding normalized bac duration ($\tau_u/\tau$), see Fig.~4.B. Specifically, in each trial (5 trials on each terrain), we programmed the robot to run for two periods such that there was at least one bac generated by each leg (6 legs visualizable from side-view camera). The correlation in these quantities indicates that bac contamination is an important source of locomotor speed variability, in accord with our model assumption. Further, we measured the distributions of average $v/v_{\text{open}}$ with different combinations of temporal and spatial redundancy (Fig.~4.D). Specifically, we measured the average displacement (over $T$ cycles, $T$ is specified in Fig.~4.D) of a 6-legged and a 12-legged robot on each terrain. The measured cumulative distribution function of average $v/v_{\text{open}}$ were in qualitative agreement with predictions from numerical tests (Fig.~3.E). The sources for discrepancies between numerical and experimental tests include over-simplification of terrain characterization and contact error from robot motor noise.

As discussed earlier, for a given transport distance $D$, the average and the variation of transport duration are both important metrics to evaluate matter transport performance. We measure the transport duration (in units of numbers of periods) as a function of $D$ for a 6-legged robot and a 14-legged robot over a rugose terrain ($R_g=0.32$). Similar to our numerical prediction in Fig.~3.F, experimental results also show that spatial redundancy significantly reduces the variation (illustrated by error bars) in transport duration (Fig.~4.C). Thus, we can guarantee predictable/reliable speed on a noisy terrain during open-loop non-inertial matter transport, analogous to that of the predictability of inertia-driven dynamics on noiseless (e.g., rails, roads) terrain.

To determine how spatial redundancy affects transport duration, we recorded $T_{[D=60]}$, the time required for a robot to locomote 60 cm over rugose terrains, for robots with varying number of legs on $R_g=\{0,\ 0.17,\ 0.32\}$ (Fig.~4.E, 10 trials in each condition). A hexapod can eventually self-transport 60 cm but there is a large variation in $T_{[D=60]}$. In contrast, systems with 16 legs can finish the self-transport task with short average time and small variation. 
Moreover, for systems with sufficiently high spatial redundancy (e.g., $N\geq5$), further increases in $N$ do not result in significant changes in $T_{[D=60]}$, including both the average and variation. This finding is consistent with our numerical prediction of the marginal benefit from spatial redundancy (Fig.~3.G).

From the Shannon scheme for signal transmission, it is reasonable to anticipate improved performance with more elaborate coding schemes, which we define as designing the terradynamic interaction profile of bacs (e.g., instantaneous thrust function $f(t)$ and thrust-velocity relationship). One straightforward method to modulate bac interaction profile is to change the body wave amplitude~\cite{chong2023self} or number of waves on the body~\cite{chong2021frequency} which alters the temporal and spatial distribution of bacs and the associated body postures. To illustrate the potential of appropriate coding, we show one example of gait design modulation. Specifically, we imposed a vertical wave along the body such that the duration of bacs ($\tau$) is actively and systematically shortened (contact modulation, see SM Section 1.5). We tested the performance of contact-modulated (CM) multi-legged robots over our rugose terrains and observed improved locomotion robustness over terrain rugosity (indicated by smaller error bars), although with some reduction of the nominal velocity $v_{\text{open}}$ (Fig.~4.E in the purple rectangular dashed box). Further, using sufficient spatial redundancy ($N=6$) as well as contact modulation, our multi-legged robot is capable of traversing diverse laboratory (obstacles, inclines and walls) environments~(Fig.~1.B and in Movie S1) and field-like environments (granular media, pebbles, and rock piles) with completely open-loop operations.

\section*{Discussion}

One value of our framework lies in its codification of the benefits of redundancy which lead to locomotor robustness to environmental contact ``errors" without requiring sensing. This is in contrast to the prevailing paradigm of contact error prevention in the conventional sensor-based closed-loop controls which take advantage of visual, tactile, or joint torque information from the environment to change the robot dynamics~\cite{raibert2008bigdog,hutter2016anymal}. In this way, the complexity of matter transport can be transferred from the real-time feedback-based control (e.g., dealing with the flow of sensor information) to pre-programmed gait design. Thus, our framework simplifies matter transport tasks such as search-and-rescue~\cite{whitman2018snake}, extraterrestrial exploration~\cite{naclerio2021controlling} or even micro-robotics~\cite{abbott2007robotics}, where robot deployments are often preferred yet challenging due to unpredictable terradynamic interactions and unreliable sensors. 

However, sensory feedback can clearly be of value when a robot becomes "stuck" or when terradynamic interactions vary significantly (e.g. moving from low to high rugosity terrain). In such cases, the robot can understand its state (via proprioception or exteroception)~\cite{borenstein1997mobile} and change dynamics accordingly. An analogy from information theory, error-detection coding facilities augmentation of our framework with sensory-based control: in error-detection coding the presence of a ``reverse channel" can facilitate the re-transmission of signals, thereby improving signal transmission accuracy~\cite{shannon1948mathematical,pierce2012introduction}. The introduction of feedback can increase the capacity of a noisy channel and therefore decrease the coding complexity~\cite{tatikonda2008capacity}. We posit that a sensor-based framework in locomotion shares a mechanism similar to error-detection coding. When the locomotion system is already equipped with redundant bacs (e.g., many legs), the benefit of adding more bacs can be marginal. In these cases, sensors (e.g., monitoring foot contact) detecting bac contamination could facilitate rapid environmental adaptation (e.g., whole-body gait adaptation or local leg placement adaptation) and further improve locomotion performance. Thus, a combination of redundancy-based and sensor-based mechanisms can offer unique advantages over challenging terrains, a feature similar to hybrid automatic repeat request method (``hybrid ARQ") in signal transmission~\cite{lott2007hybrid}.

In addition to the importance of locomotion in artificial locomotors, we posit that our matter transport framework can give insights into aspects of neuromechanical and morphological evolution~\cite{chong2022coordinating} from a physics of living systems perspective. That is, animals ranging from those which generate propulsion via a single bac pair (i.e., bipeds)~\cite{marigold2008role,krasovsky2014effects} to those which utilize many bacs (i.e., myriapods)~\cite{diaz2023active} are capable of traversing complex natural terrains. The importance of environmental awareness and whole body coordination is hypothesized to diminish as the number of bacs (redundancy) increases~\cite{diaz2023active,sponberg2008neuromechanical,blasing2004mechanisms}. Thus, in biological terrestrial locomotors, there appears to be a shift towards either advanced neuromechanical control with reduced body appendages, or redundant body appendages with simplified neuromechanical control. Integration of our framework with advances in biological experimentation could yield insights into the benefits and trade-off of diverse control architectures.

\section*{Acknowledgments}
We thank Prof. Todd Murphey, Prof. Paul Umbanhowar, Prof. Guillaume Adrien Sartoretti, Dr. Di Luo, Thomas Berrueta, and Xiaoxiao Sun for helpful discussion. We thank Kelimar Diaz for proofreading.
\textbf{Funding:} The authors are grateful for funding from NSF-Simons Southeast Center for Mathematics and Biology (Simons Foundation SFARI 594594), Georgia Research Alliance (GRA.VL22.B12), Army Research Office (ARO) MURI program, Army Research Office Grant W911NF-11-1-0514 and a Dunn Family Professorship. \textbf{Author contributions:} B.C. designed the study/model and performed and numerical simulations. J.H. collected the raw data for the robophysical experiments. J.H., T.W., D.S., and B.C. built the experimental platform. D.I., G.B., and B.C. performed the mathematical proof. D.I.G. oversaw the study. All authors contributed to the preparation of the manuscript and were involved in the interpretation of results. We believe that some of the subject matter herein may be implicated in one or more pending patent applications such as, for example PCT Patent Application No. PCT/US2022/043362, entitled “DEVICES AND SYSTEMS FOR LOCOMOTING DIVERSE TERRAIN AND METHODS OF USE”, which claims the benefit of priority to U.S. Provisional App. No. 63/243,435 filed 09/13/2021, and U.S. Provisional App. No. 63/318,868 and filed 03/11/2022. \textbf{Competing interests:} The
authors declare that they have no competing interests. \textbf{Data and materials availability:} All data that support the claims in this manuscript are available on the Zenodo repository (46).

\section*{List of supplementary content }
Materials and Methods\\
Supplementary Text\\
Figs. S1 to S6\\
Movie S1 caption\\
References \textit{(47-48)}

\clearpage
\begin{figure*}[ht]
\centering
    \includegraphics[width=1\linewidth]{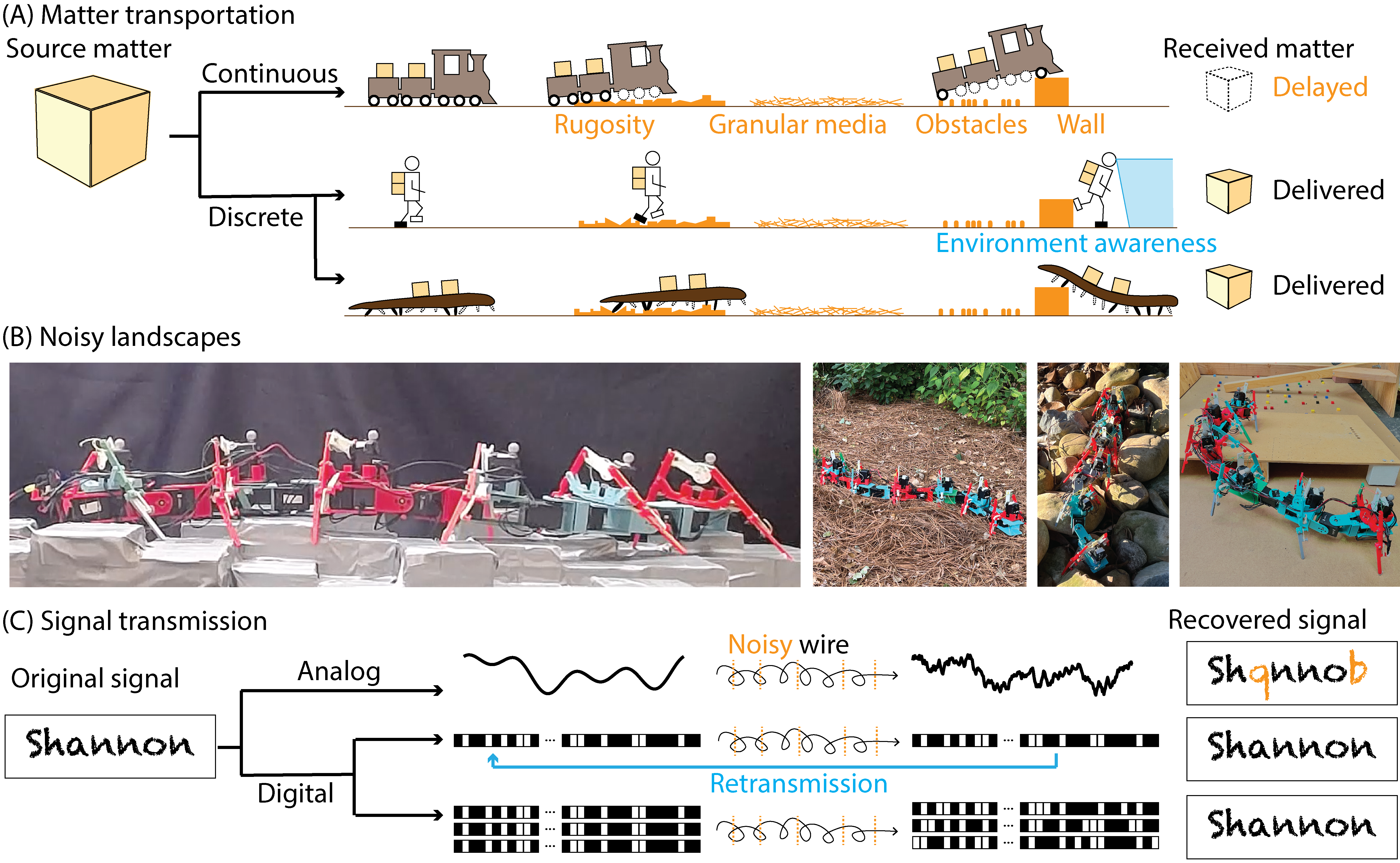}
    \caption{\textbf{Signal transmission and matter transport.} (A) Matter transport with both continuous and discrete active contacts can be effective on noise-free tracks. Discrete redundant contacts enable effective matter transport over rugose tracks via redundancy or environmental awareness. (B) Multi-segmented  robophysical locomotors with directional-compliant legs traverse noisy landscapes: (from left to right) a laboratory model of rugose terrain, entangled granular media, pebbles, and stairs (slopes). (C) The signal transmission paradigm. We compare analog and digital signal transmission through noisy wires. A digital signal allows reliable transmission through a noisy wire via either redundancy or a re-transmission channel.}

\end{figure*}
\clearpage
\begin{figure*}[ht!]
    \centering
    \includegraphics[width=1\linewidth]{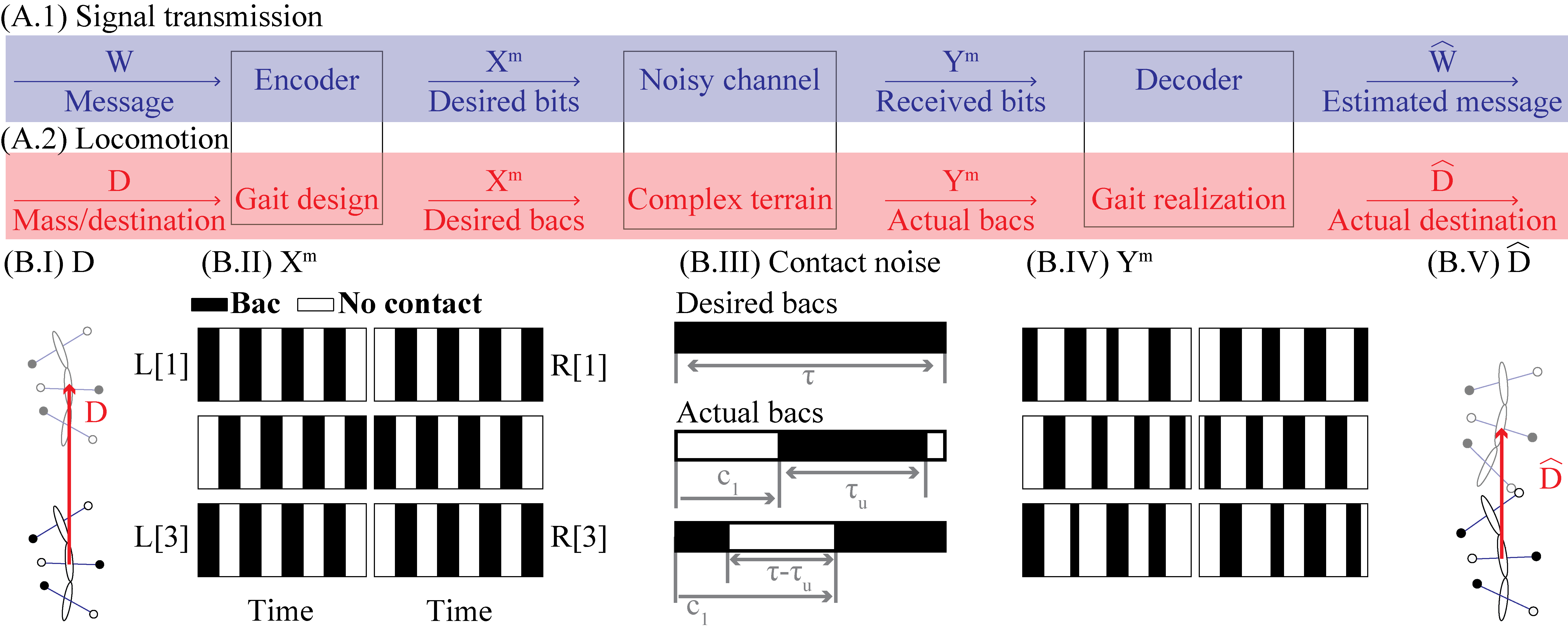}
    \caption{\textbf{Framing the matter transport problem as a sequence of basic active contacts (bacs).} (A) The correspondence of processes in (A.1) signal transmission (adapted from~\cite{shannon1948mathematical}) and (A.2) locomotion (matter transport). (B.I) A multi-legged robot, the matter to be transported to a destination $D$. (B.II) The desired bac sequence to reach the locomotion destination. (B.III) Noisy landscapes can introduce contact errors such as delaying bacs and shortening the duration of bacs. We compare the desired bac (which spans a duration $\tau$) and two terrain-contaminated bacs (each begins at $c_1$) with shorter duration ($\tau_u$). (B.IV) A bac sequence contaminated by contact errors leads to a (B.V) locomotion destination $\hat{D}$ smaller than the expected $D$.}
\end{figure*}

\clearpage
\begin{figure}[t!]
    \centering
    \includegraphics[width=0.5\linewidth]{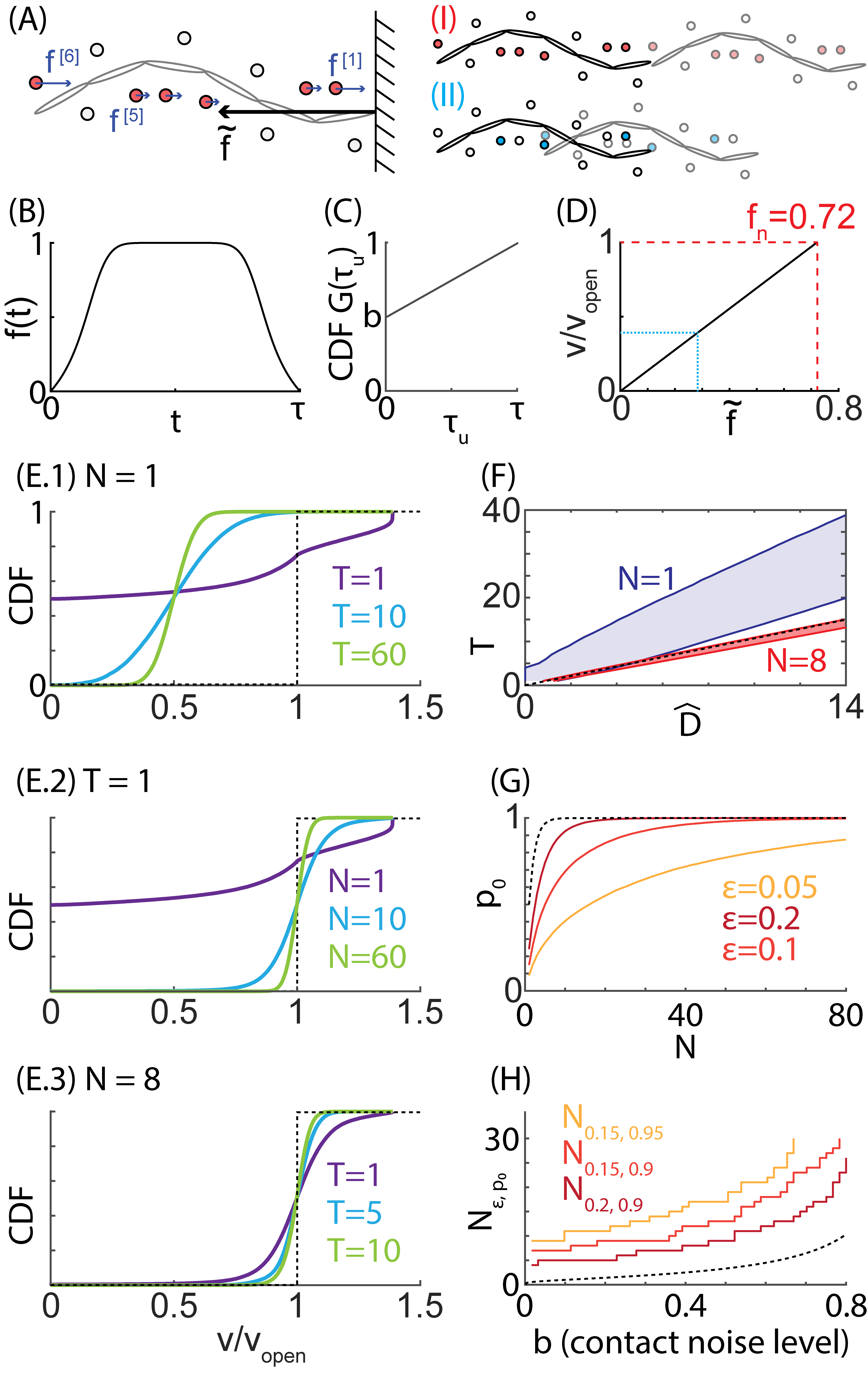}
    \caption{\textbf{Numerical simulation to test the matter transport framework.} (A) Illustrations of (\textit{Left}) thrust generation from bacs and (\textit{Right}) thrust-velocity relationship. Self-propulsion with (I) nominal contact and (II) contact errors are compared. (B) The instantaneous thrust $f(t)$ as a function of time, derived from~\cite{chong2023self}. (C) The cumulative distribution function (CDF) of $\tau_u$. (D) The thrust-velocity (normalized by the nominal velocity) relationship. (E) The numerical CDF of terrain-disturbed velocity for robots with different combinations of temporal ($T$) and spatial ($N$) redundancy. In all sub-panels, black dashed curves denote the nominal CDF. (F) Numerically calculated 90\% CI of actual destination $\hat{D}$ as a function of $T$. We compared two spatial redundancies: $N=1$ (blue) and $N=8$ (red). The nominal $\hat{D}$-$T$ relationship is shown in black dashed curve. (G) The probability of successful scheduled matter transport ($p_0$) as a function of spatial redundancy evaluated at different tolerance $\epsilon$. We plot the theoretical predicted limit in black dashed curve. (H) The relationship between $N_{\epsilon, p_0}$, the minimal spatial redundancy required to achieve the desired success probability ($p_0$) subject to tolerance $\epsilon$, and $b$, contact noise level. The black dashed line denotes the theoretical predicted limit when $p_0=0.9$.}
\end{figure}

\clearpage
\begin{figure}[t!]
    \centering
    \includegraphics[width=0.5\linewidth]{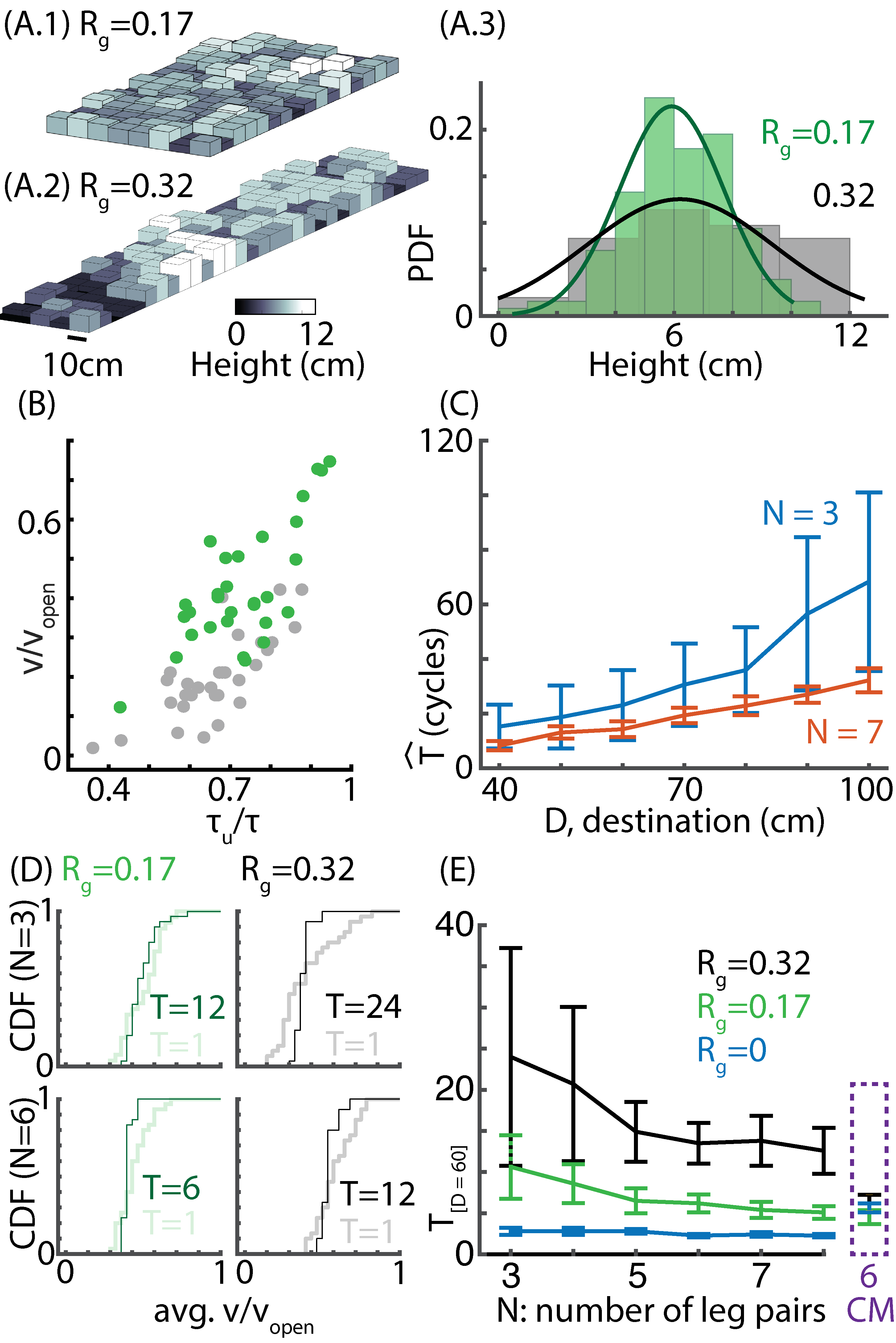}
    \caption{\textbf{Experimental test of the matter transport framework on multi-legged robots.} (A) Renderings of rugose terrains with rugosity (A.1) $R_g=0.17$ and (A.2) $R_g = 0.32$. The block height distribution is shown on the right panel. (B) Bac contamination leads to speed degradation in a 12-legged robot on rugose terrains. Each point denotes the robot's normalized inter-step averaged velocity  $v/v_{\text{open}}$) and a corresponding contact duration ($\tau_u/\tau$) in one bac. Color schemes are identical to (A). (C) The empirical transport duration ($\hat{T}$, units: gait periods) as a function of destination distance $D$. We compare two robots with 14 (red) and 6 (blue) legs. There is a large variation of transport duration for $6$-legged robot, and the variation grows as destination distance increases. 14-legged robot has a comparably tightly bounded transport duration. (D) The empirical distribution of average velocity on terrains with (\textit{left}) $R_g=0.17$ and (\textit{right}) $R_g = 0.32$ on (\textit{Top}) the 6-legged robot and (\textit{Bottom}) the 12-legged robot. The empirical distributions were obtained from 30 trials. (E) For robot with different numbers of leg pairs $N$, we recorded $T_{[D=60]}$, the number of periods required to transport $D=60$cm on terrains with (blue) $R_g=0$, (green) $R_g=0.17$, and (black) $R_g=0.32$. The error bar was calculated from at least 10 trials. $T_{[D=60]}$ for contact modulated gaits are illustrated in the purple rectangle. } 
\end{figure}

\clearpage

\end{document}